\def\year{2021}\relax
\documentclass[letterpaper]{article} 
\usepackage{aaai21}  
\usepackage{times}  
\usepackage{helvet} 
\usepackage{courier}  
\usepackage[hyphens]{url}  
\usepackage{graphicx} 
\def\UrlFont{\rm}  
\usepackage{natbib}  
\usepackage{caption} 
\title{Reducing ReLU Count for Privacy-Preserving CNN Speedup}

\author {
    Inbar Helbitz,\textsuperscript{\rm 1}
    Shai Avidan \textsuperscript{\rm 2} \\
}
\affiliations {
    \textsuperscript{\rm 1} School of CS, Tel-Aviv University \\
    \textsuperscript{\rm 2} School of EE, Tel-Aviv University \\
    inbarhelbitz@mail.tau.ac.il, avidan@eng.tau.ac.il
}

\begin{document}

\maketitle

\begin{abstract}
Privacy-Preserving Machine Learning algorithms must balance classification accuracy with data privacy. This can be done using a combination of cryptographic and machine learning tools such as Convolutional Neural Networks (CNN). CNNs typically consist of two types of operations: a convolutional or linear layer, followed by a non-linear function such as ReLU. Each of these types can be implemented efficiently using a different cryptographic tool. But these tools require different representations and switching between them is time-consuming and expensive. Recent research suggests that ReLU is responsible for most of the communication bandwidth. 

ReLU is usually applied at each pixel (or activation) location, which is quite expensive. We propose to share ReLU operations. Specifically, the ReLU decision of one activation can be used by others, and we explore different ways to group activations and different ways to determine the ReLU for such a group of activations.

Experiments on several datasets reveal that we can cut the number of ReLU operations by up to three orders of magnitude and, as a result, cut the communication bandwidth by more than $50\%$.
\end{abstract}


\section{Introduction}

Deep Learning algorithms have made great progress in recent years. Therein lies the problem. In order to classify an image, a Deep Learning algorithm must be exposed to its content. Yet, an image contains private information that should not be exposed.
Protecting the privacy of the data is paramount in the medical, financial, and military domains, to name a few.

Privacy-Preserving Machine Learning algorithms are designed to address this problem. Such an algorithm will protect the privacy of the data, as well as that of the network, while performing the task at hand. For example, assume that Alice owns an image and Bob owns a Convolutional Neural Network (CNN) for image classification. Their goal is to collaborate in classifying the image without revealing any additional information. Alice should learn nothing about the weights of Bob's network, while Bob should learn nothing about the content of Alice's image. Only the classification result should be revealed to the agreed upon party.

This problem can be solved using cryptographic tools. The leading approach in recent work is to use secure Multi-Party Computation (MPC) techniques. In this setting, Alice and Bob securely evaluate the network layer by layer until an outcome is reached. At each stage, Alice and Bob hold private shares of the data (or the intermediate results of the network so far). Then, they use Homomorphic Encryption to compute the convolutional and linear component. The non-linear ReLU operation requires a different representation and so Alice and Bob must switch their data representation to compute the ReLU outcome of their operation. And the process repeats for the next layer. Switching to and from ReLU representation is time-consuming and requires considerable communication bandwidth.

The goal of this paper is to develop an algorithm that reduces the ReLU count in a given CNN. Our solution is based on {\em ReLU Sharing} -  a group of pixels (or activation units down the network) operate based on a single ReLU operation. Today, each activation performs the ReLU operation on the outcome of the convolution at the activation's site. In ReLU sharing, the ReLU decision at one activation location is shared by others. For example, if we assume that nearby activations are highly correlated, then it is reasonable to assume that their ReLU result will be similar. Therefore, it might be enough to perform ReLU on one activation, and use that result for other activations in its neighborhood.

We consider different methods to group activations and evaluate different ways for a group of activations to perform a shared ReLU operation. We evaluate our algorithm on several data sets and report the results. We find that in some cases we can reduce the ReLU count by up to three orders of magnitude in early layers of the network and cut the total number of ReLU operations by nearly $80\%$ with less than a $2\%$ drop in accuracy.

\section{Background}

Secure Multi Party Computation (MPC) was first investigated in Cryptography. Yao~\cite{Yao82} proposed a Garbled Circuit (GC) to solve the Millionaire Problem: Two Millionaires, Alice and Bob, want to determine which one has more money, without revealing their true wealth. Yao reduced the problem to a circuit that is evaluated jointly and securely by both parties. Observe that the Millionaire Problem is the same as ReLU. 

Since every computer program can be reduced to a Boolean circuit, then any computer program can be evaluated securely. Later, Goldreich {\em et al.} \cite{GMW87} extended MPC to more than two parties. In practice, GC suffers from two main drawbacks. It requires multiple rounds of communication and is extremely slow to compute.

Alternatively, one can use Fully Homomorphic Encryption (FHE), which is an encryption scheme that supports addition and multiplication of encrypted numbers. This allows Alice to encrypt her image and send it to Bob. Bob, in turn, can run his Convolutional Neural Network (CNN) on the encrypted data and return the result. Because FHE does not support a non-linearity such as ReLU, Bob will have to use an alternative non-linearity such as a low degree polynomial or {\em tanh}.

This elegant solution guarantees the privacy of both parties, while minimizing communication bandwidth. Alice simply sends her encrypted image to Bob and gets in return its classification. This solution was adopted in CryptoNets~\cite{CryptoNets}.
Unfortunately, FHE is painfully slow and running it on real images is not feasible currently. 

Which leads us to present day solutions. Instead of relying on FHE, most methods today evaluate the CNN layer by layer using a combination of GC and Homomorphic Encryption (HE). These algorithms require several rounds of communication and rely on Homomorphic Encryption (HE), which is less demanding than FHE. Specifically, HE supports either addition or multiplication of encrypted data and is much faster to compute. This means that Bob can run the convolutional layers of his network on Alices's encrypted data, but not the ReLU non-linearity. The ReLU operation can be implemented using a Garbled Circuit. Switching between HE and GC is very time consuming and requires several rounds of communication between Alice and Bob.

Most of the recent progress in Secure Deep Learning mixes and matches these cryptographic tools. DeepSecure \cite{DeepSecure} build on top of Yao's Garbled Circuit a system for secure Deep Learning, and test it on several data sets from different domains. In the context of Vision, they report results on the MNIST dataset.

SecureML~\cite{SecureML} use two-party computation to address the problem. In particular, they develop new techniques to support secure arithmetic operations on shared decimal numbers, and propose MPC-friendly alternatives to non-linear
functions such as sigmoid and softmax that are superior to prior work.

MiniONN~\cite{MiniONN} transforms a CNN into an Oblivious CNN. They detail how each component in a CNN can be implemented obliviously and show that some of the computational burden can be shifted to an offline stage. They show considerable improvements over both CryptoNets and SecureML. GAZELLE~\cite{Gazelle} further improved the efficiency of the linear layer computation.

Chamelon~\cite{Chameleon} combines secure function evaluation with secret sharing for the linear layers and MPC or GC for the nonlinear operations. They report results that are two orders of magnitude faster than CryptoNets~\cite{CryptoNets} and about four times better than~\cite{MiniONN}.

Recently, FALCON~\cite{FALCON} suggested the use of FFT to accelerate the secure computation of the linear layer. They are also the first to introduce a secure softmax evaluation.

The work most closely related to us is that of Shafran {\em et al.}~\cite{Shafran2020}. They, too, observe that ReLUs are expensive to compute and propose several ways to reduce their count.

\section{Method}

Our goal is to reduce the number of ReLU operations, as this is a major factor in the cost of secure inference. Our key insight is that activations are highly correlated with each other and hence the ReLU decision of one can be used for another. The question is how to share ReLU decisions without compromising privacy?

Let ${\bf x} \in R^{n \times m}$ denote an $n \times m$, single channel image, and let ${\bf w} \in R^{k \times k}$ denote a convolution kernel. The convolution result is given by ${\bf s} = {\bf x} * {\bf w}$ and the ReLU on top of it is ${\bf y} = \mbox{ReLU}({\bf s})$, which is defined element-wise as:

\begin{equation}
    \mbox{ReLU}({\bf s(p)}) = \left\{ \begin{array}{cc}
                    {\bf s(p)} & {\bf s(p)} \geq 0\\
                    0 & \mbox{otherwise} \\
                    \end{array}
                    \right.
\end{equation} 
where ${\bf p}$ is (2D) activation location.

We can generalize ReLU such that a ReLU decision of a group of activations is then being used by each activation independently. Formally:
\begin{equation}
    \mbox{gReLU}({\bf s(p)}, {\bf v}^T{\bf s}) = \left\{ \begin{array}{cc}
                                                {\bf s(p)} & {\bf v}^T{\bf s} \geq 0\\
                                                0 & \mbox{otherwise}\\
                                                \end{array}
                                                \right.
\end{equation} 
where ${\bf v}$ is a weight vector, the size of ${\bf s}$, and we treat them both as column vector for ease of notation. If, for ${\bf s(p)}$, we take ${\bf v}$ to be a one hot vector with ${\bf v(p)}=1$ then we are back to standard ReLU. If, on the other hand, we take ${\bf v}$ to be a one hot vector with ${\bf v(q)}=1$ then the ReLU decision for pixel ${\bf p}$ is determined by pixel ${\bf q}$. 

Of course, ${\bf v}$ does not have to be a one hot vector, and we do not have to use just a single vector ${\bf v}$. A straightforward generalization is to group the activations in the channel plane into patches, assign a different weight vector ${\bf v_i}$ to each patch and learn the weights.

\begin{equation}
    \mbox{gReLU}({\bf s(p)}, {\bf v_i}^T{\bf s}) = \left\{ \begin{array}{cc}
                    {\bf s(p)} & {\bf v_i}^T{\bf s} \geq 0 \land {\bf p} \in i\\
                    0 & \mbox{otherwise}\\
                    \end{array}
                    \right.
\end{equation} 
where ${\bf p} \in i$ is a shorthand for activation ${\bf p}$ belongs to patch $i$.

\begin{figure}
\begin{center}
  \includegraphics[width=1.0\linewidth]{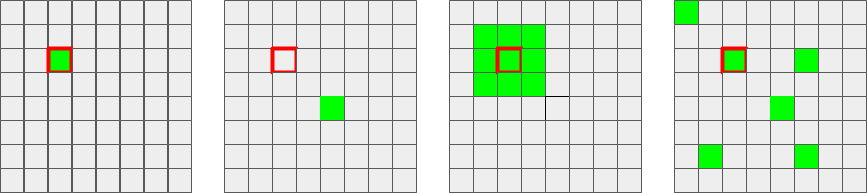}
\end{center}
  \caption{Illustration of ReLU versions. Four toy $8 \times 8$ pixel images. The green pixels determine the ReLU of the pixel with red boundary. From left to right: Regular ReLU, gReLU - the ReLU of the red pixel is determined by a different pixel (marked in green), gReLU - the ReLU of the red pixel is determined by all green pixels (grouped based on proximity in image plane), and gReLU - the ReLU of the red pixel is determined by a group of green pixels (clustered based on training data).}
\label{fig:ReLUCorr}
\end{figure}

Figure~\ref{fig:ReLUCorr} shows the different ReLU variants that can be used. From left to right we see, first, a regular ReLU, where the ReLU operation is applied directly to the activation itself. Next we show gReLU, where the ReLU decision of activation ${\bf q}$ (green activation) is applied to the red-boundary activation ${\bf p}$. The last two sub-plots show two different grouping techniques that can be used. In both cases, the gReLU decision of the weighted sum of the green activations is used to determine the ReLU operation of the red-boundary activation.

The first clustering technique is oblivious to data and simply breaks the channel into small patches (i.e., $3 \times 3$ patches). The second clustering technique we consider is based on the training data, as we explain next.

Assume we have $N$ training sample and let ${\bf y^j({\bf p})}$ be a binary vector that denotes the ReLU decision of activation ${\bf p}$ for example $j$. Let ${\bf y({\bf p})} = ( y^1({\bf p}), \cdots, y^N({\bf p}) )$ denote an $N$ dimensional vector of the ReLU response, at pixel ${\bf p}$ of all training examples. The number of such vectors is the number of activations in the channel, which is $nm$. We now cluster these vectors to find activations that are correlated with each other.

\begin{figure}
\begin{center}
  \begin{minipage}[b]{0.15\textwidth}
      \centerline{\includegraphics[width=1\textwidth]{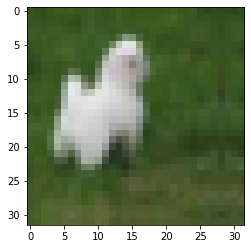}}
  
  \end{minipage}\hfill
  \begin{minipage}[b]{0.15\textwidth}
      \centerline{\includegraphics[width=1\textwidth]{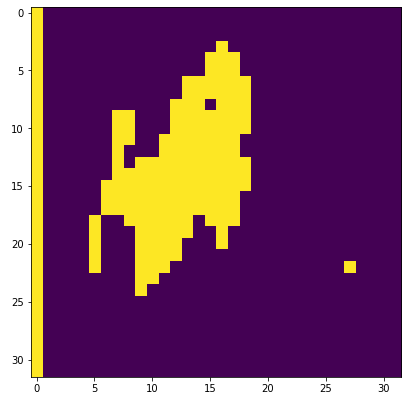}}
  
  \end{minipage}\hfill
  \begin{minipage}[b]{0.15\textwidth}
      \centerline{\includegraphics[width=1\textwidth]{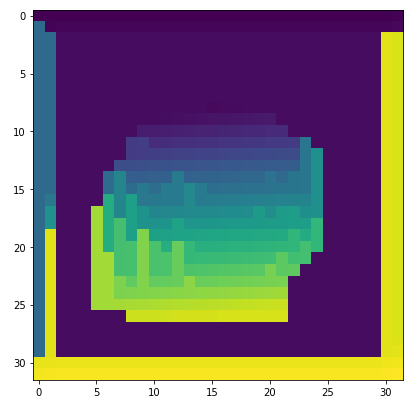}}
  
  \end{minipage}
\end{center}
  \caption{ReLU clustering. (left) An example input image. (Center) ReLU activation map for that image for one of the channels in the first layer. (Right) A pixel clustering map based on the ReLU activation maps of all images in the training set.}
\label{fig:ReLUClusterIllustration}
\end{figure}

Figure~\ref{fig:ReLUClusterIllustration} shows an example taken from the CIFAR-10 dataset. On the left we see an example input image, at the center we see its corresponding ReLU activation map ${\bf y^j}$ for one particular channel in the first layer of the network and on the right we see the clustering map. There is a different weight vector ${\bf v_j}$ for each color in the clustering map (not shown). The weights are set during training.

One problem with the gReLU operator is that all pixels in the group behave the same. Either they all maintain their input value, or they are all set to zero. If we use a single weight vector per channel, this means that either the entire channel is preserved or zeroed out. To avoid this situation, we introduce a noisy version of gReLU:
\begin{equation}
    \mbox{ngReLU}({\bf s_j}, {\bf v}^T{\bf s}, p) = \left\{ \begin{array}{cc}
                    \mbox{gReLU}({\bf s_j}, {\bf v}^T{\bf s})  & \mbox{Ber}(p)\\
                    \mbox{gReLU}({\bf s_j}, -{\bf v}^T{\bf s}) & \mbox{otherwise}\\
                    \end{array}
                    \right.
\end{equation} 
where $\mbox{Ber}(p)$ is the Bernoulli distribution with parameter $p$. This way, some activations maintain their value in case the ReLU decision is to zero out. In case the ReLU decision was to keep the activation value, then zero out some of the activations is akin to Dropout, which is a known technique in the literature.

In terms of security, the first clustering method, which is oblivious to the data, preserves the privacy of the data. No additional information is leaked to either party, other than the fact that some activation units share the same ReLU output, which is akin to sharing the overall architecture of the network between the parties. The second clustering method, which is based on the {\em training} data, reveals information, but only about the training data and {\em not} about the query image at inference time. See, for example, Figure~\ref{fig:ReLUClusterIllustration} or Figure~\ref{fig:ReLUCluster}.

\section{Experiments}

\paragraph{Datasets}
We test our algorithms on the CIFAR-10, SVHN and Fashion-MNIST datasets. They are 10-class classification problems that take as an input an RGB images of size $32 \times 32$ pixels (CIFAR-10, SVHN) or a grayscale image of size $28 \times 28$ (Fashion-MNIST).

\paragraph{Implementation Details}
In order to evaluate running times, number of communication rounds and the communication bandwidth in the encrypted settings we used the PySyft library \cite{PySyft}, a framework for secure and private deep learning within the PyTorch framework. PySyft is a 3-party computation settings implementing the secureNN protocol \cite{SecureNN}.

Table~\ref{table:securenn} gives the round complexity and communication bandwidth of the different operations involved in computing a single layer in a CNN for the secureNN protocol. $Linear_{m,n,v}$ denotes a matrix multiplication of dimension $m \times n$ with $n \times v$. $Conv2d_{m,i,f,o}$ denotes a convolutional layer with input $m \times m$, i input channels, a filter of size $f \times f$, and o output channels. DReLU denotes the binary calculation if a value is greater than 0 for a single value. Mul denotes a single value multiplication. A single ReLU operation is composed of a DReLU and a Mul operations. The round complexity of a convolution layer is 2, while the total round complexity of ReLU layer is 10.

All communication is measured for $l$-bit inputs (64-bit for PySyft) and $p$ denotes the field size (67 for PySyft), for more details see \cite{SecureNN}. ReLU clearly dominates the process.

\begin{table}
\small
\begin{center}
\begin{tabular}{|l|c|l|}
\hline
Protocol & Rounds & Communication \\
\hline\hline
$Linear_{m,n,v}$ & $2$ & $(2mn + 2nv + mv)l$ \\
$Conv2d_{m,i,f,o}$ & $2$ & $(2m^2f^2i+2f^2oi+m^2o)l$ \\
DReLU & $8$ & $(8 logp + 19)l \approx 83l$ \\
Mul & $2$ & $5l$ \\
\hline
\end{tabular}
\end{center}
\caption{Round and communication complexity for secureNN protocols. Linear and convolutional layers require just 2 round of communication. ReLU operation requires both DReLU and Mul operations, totaling in 10 rounds of communication. See text for further details.}
\label{table:securenn}
\end{table}

We evaluate the run times of our models on a single Intel Core i5 CPU and 8GB RAM, which lets us evaluate improvement in performance. In our setting both the model and the data are shared between 2 parties (Alice and Bob), the third party doesn’t hold shares but helps with common randomness and computations. In our setting all 3 parties live on the same machine and do not communicate over the network, so all running times do not consider network latency.

\paragraph{CIFAR-10} For the CIFAR-10 dataset, we use a CNN that is based on the architecture proposed in MiniONN~\cite{MiniONN}. See table~\ref{table:cifar_cnn}. The network consists of a sequence of convolutions, ReLU, max pooling, and fully connected layers. We replaced the max pooling operation with average pooling, to reduce ReLU count. The number of ReLU operations in the first and second layers account for about $75\%$ of all ReLU operations, so we focus on them.

\paragraph{SVHN and Fashion-MNIST} For the SVHN and the Fashion-MNIST datasets, we use the CNNs in tables~\ref{table:svhn_cnn} and~\ref{table:fashion_cnn}, respectively. Since they are small networks we focus on the first ReLU layer of the networks. For SVHN the first layer accounts for 92\% of all ReLU operations, and for Fashion-MNIST the first layer accounts for 60\%.


\begin{table}
\small
\begin{center}
\begin{tabular}{|l|c|c|}
\hline
Operation & patch size & Output size \\
\hline\hline
conv & $3 \times 3$ & $32 \times 32 \times 64$ \\
conv & $3 \times 3$ & $32 \times 32 \times 64$ \\
Avg Pool & $2 \times 2$ & $16 \times 16 \times 64$ \\
conv & $3 \times 3$ & $16 \times 16 \times 64$ \\
conv & $3 \times 3$ & $16 \times 16 \times 64$ \\
Avg Pool & $2 \times 2$ & $8 \times 8 \times 64$ \\
conv & $3 \times 3$ & $8 \times 8 \times 64$ \\
conv & $1 \times 1$ & $8 \times 8 \times 64$ \\
conv & $1 \times 1$ & $8 \times 8 \times 16$ \\
FC   & $1024 \times 10$ & $1 \times 10$ \\
\hline
\end{tabular}
\end{center}
\caption{CIFAR-10 Architecture. The input is a $32 \times 32 \times 3$ RGB image. After each convolution we perform regular ReLU. The first two layers account for about $75\%$ of the ReLU operations in the network.}
\label{table:cifar_cnn}
\end{table}

\begin{table}
\small
\begin{center}
\begin{tabular}{|l|c|c|}
\hline
Operation & patch size & Output size \\
\hline\hline
conv & $4 \times 4$ & $29 \times 29 \times 32$ \\
Avg Pool & $3 \times 3$ & $9 \times 9 \times 32$ \\
conv & $4 \times 4$ & $6 \times 6 \times 64$ \\
FC   & $2304 \times 10$ & $1 \times 10$ \\
\hline
\end{tabular}
\end{center}
\caption{SVHN Architecture. The input is a $32 \times 32 \times 3$ RGB image. After each convolution we perform regular ReLU. The first ReLU layer accounts for about $92\%$ of the ReLU operations in the network.}
\label{table:svhn_cnn}
\end{table}

\begin{table}
\small
\begin{center}
\begin{tabular}{|l|c|c|}
\hline
Operation & patch size & Output size \\
\hline\hline
conv & $3 \times 3$ & $28 \times 28 \times 32$ \\
Avg Pool & $2 \times 2$ & $14 \times 14 \times 32$ \\
conv & $3 \times 3$ & $14 \times 14 \times 64$ \\
Avg Pool & $2 \times 2$ & $7 \times 7 \times 64$ \\
Dropout 0.5   & $-$ & $-$ \\
FC   & $3136 \times 128$ & $1 \times 128$ \\
FC   & $128 \times 10$ & $1 \times 10$ \\
\hline
\end{tabular}
\end{center}
\caption{Fashion-MNIST Architecture. The input is a $28 \times 28$ grayscale image. After each convolution we perform regular ReLU. The first ReLU layer accounts for about $60\%$ of the ReLU operations in the network.}
\label{table:fashion_cnn}
\end{table}

\paragraph{Reducing ReLU Count} The first set of experiments is designed to measure the impact of reducing ReLU count on the CNN accuracy. This can be achieved by either clustering activations into small image patches (i.e., data agnostic), or clustering together based on their behaviour on the training data (i.e., data dependent).

We start with sharing ReLUs over small patches of activations (i.e., pixels in the first layer). In particular, we evaluate non-overlapping patches of size $3 \times 3$ and $4 \times 4$ activations, and perform this on both the first and second layer. For comparison, we also evaluate the two extreme cases. The first is the regular CNN practice (i.e., $1 \times 1$ patches), the other is the extreme case where the entire image (i.e., patch size of $32 \times 32$) is based on a single ReLU.

In each case we assign a weight vector and set its weights during training. All activations in the patch base their ReLU decision on it. This partitioning is agnostic to the content of the images. The top part of Table~\ref{table:cifar_fixed} shows the results of this experiment on the CIFAR-10 model. Observe the nice correlation between the number of ReLUs and the accuracy.

For the SVHN and Fashion-MNIST model see Table~\ref{table:svhn_fashion_fixed}. We get less than $1\%$ loss in accuracy, compared with the original accuracy of the models but with only $81$ ReLUs per channel, compared to $841$ and $784$ ReLUs per layer, respectively, in the original models.

\begin{table}
\small
\begin{center}
\begin{tabular}{|l|c|c|c|c|}
\hline
Method & Patch size & \#ReLU & Noise & Accuracy\\
\hline\hline
Original & $1 \times 1$ & $(1024,1024)$ & No & $83.59\%$\\
Uniform & $3\times3$ & $(121,121)$ & No & $82.65\%$\\
Uniform & $4\times4$ & $(64, 64)$ & No & $81.58\%$ \\
Uniform & $32\times32$ & $(1, 1)$ & No & $81.13\%$\\
\hline
Layer 1 only & $32\times32$ & $(1, 1024)$ & No & $84.36\%$\\
Layer 2 only & $32\times32$ & $(1024, 1)$ & No & $79.55\%$\\
\hline
Noise & $32\times32$ & $(1,1)$ & $20\%$ & $80.10\%$\\
FC+Noise & $32\times32$ & $(1,1)$ & $20\%$ & $82.01\%$\\
\hline
\end{tabular}
\end{center}
\caption{CIFAR-10 ReLU sharing on image patches. The expression $(a,b)$ in the \#ReLU column counts the number of ReLU operations in the first ($a$) and second ($b$) per channels. The original network requires $2048=1024+1024$ ReLU operations per channel with an accuracy of $83.59\%$. The "Layer 1 only" ("Layer 2 only") replaces all ReLU operations of the first (second) layer with a single ReLU operation. The "FC+Noise" method requires just 1 ReLU operations per channel with an accuracy of $82.01\%$. A drop of three orders of magnitude in ReLU count. See more details in the text.}
\label{table:cifar_fixed}
\end{table}

The next part of Table~\ref{table:cifar_fixed} reports results of keeping one layer intact and replacing the other layer with a single ReLU per layer. We see that performance actually increases in the case of the first layer but degrades in the case of the second layer. As we go deeper into the network we need more and more ReLU decisions in order to maximize the model’s accuracy.

The extreme case of using a single ReLU decision for the whole channel is like a binary choice whether to use this channel in the classification of the image or not, since it either leaves the activation map as is or zeroes-out the entire channel. 

As discussed earlier, if the ReLU decision is zero, then the entire patch is zeroed out, leaving no information for the next pooling operation. To battle this phenomenon, we introduce noise into the ReLU result. Specifically, in this experiment both layers use a single ReLU per channel, where the ReLU is carried out on the middle activation of the channel but we add $20\%$ noise to the ReLU decision of the second layer only. Results for CIFAR-10 are reported in Table~\ref{table:cifar_fixed} and for SVHN in Table~\ref{table:svhn_fashion_fixed}. As can be seen, adding noise helped improve the results compared to working with image patches.

Next we replaced the use of a single activation to determine the ReLU outcome with  learned weights of the activations. To do that, we added a fully connected layer and performed a dot-product with the activation map and got a single value that determines the ReLU decision. Then we flipped $20\%$ of the activation decisions and according to this we perform the ReLU on the original activation map. For CIFAR-10, results are shown in table~\ref{table:cifar_fixed}. We get less than $1\%$ loss in accuracy, compared with the original accuracy of the model but with only $1$ ReLUs per channel per layer, compared to $1024$ ReLUs per layer in the original model.

\begin{table}
\small
\begin{center}
\begin{tabular}{|l|c|c|c|c|}

\multicolumn{5}{l}{\textbf{SVHN}}\\
\hline
Method & Patch size & \#ReLU & Noise & Accuracy\\
\hline\hline
Original & $1 \times 1$ & $29 \times 29$ & No & $93.14\%$\\
Uniform & $3\times3$ & $9 \times 9$ & No & $92.33\%$\\
Uniform & $32\times32$ & $1$ & No & $90.21\%$\\
\hline
Adaptive & $32\times32$ & $256$ & No & $91.63\%$\\
\hline
Noise & $32\times32$ & $1$ & $20\%$ & $91.22\%$\\
FC+Noise & $32\times32$ & $1$ & $20\%$ & $90.81\%$\\
\hline
\multicolumn{5}{l}{\textbf{Fashion-MNIST}}\\
\hline
Method & Patch size & \#ReLU & Noise & Accuracy\\
\hline\hline
Original & $1 \times 1$ & $28 \times 28$ & No & $92.11\%$\\
Uniform & $3\times3$ & $9 \times 9$ & No & $91.11\%$\\
Uniform & $32\times32$ & $1$ & No & $90.38\%$\\
\hline
Noise & $32\times32$ & $1$ & $20\%$ & $90.31\%$\\
FC+Noise & $32\times32$ & $1$ & $20\%$ & $90.20\%$\\
\hline
\end{tabular}
\end{center}
\caption{SVHN and Fashion-MNIST ReLU sharing on image patches. The \#ReLU column counts the number of ReLU operations per channel at the first layer of the network.}
\label{table:svhn_fashion_fixed}
\end{table}

\begin{figure}[!t]
\begin{center}
  \includegraphics[width=1.0\linewidth]{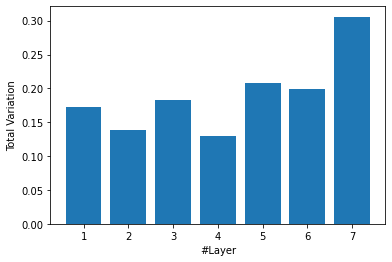}
\end{center}
  \caption{Total Variation (TV) across layers. We show the TV of the response images across the different layers of the network for the CIDAR-10 dataset. As can be seen, the TV increases with depth. We believe the two dips (at $x=2, x=4$) are because they are before an average pooling operation.}
\label{fig:ReLUTV}
\end{figure}

We observe that it is easier to compress the earlier layers of the network and conjecture that this is because the resulting image is smoother. To validate this assumption, we compute the Total Variation (the sum of absolute values of the spatial gradients in the image plane) along the layers. See Figure~\ref{fig:ReLUTV}. As can be seen, the Total Variation of the activations increases as we move deeper into the network. This means the activation map is less smooth and therefore the correlation between nearby activations is lower. As a result, we need to increase the ReLU count.

\begin{figure}
\begin{center}
  \includegraphics[width=1.0\linewidth]{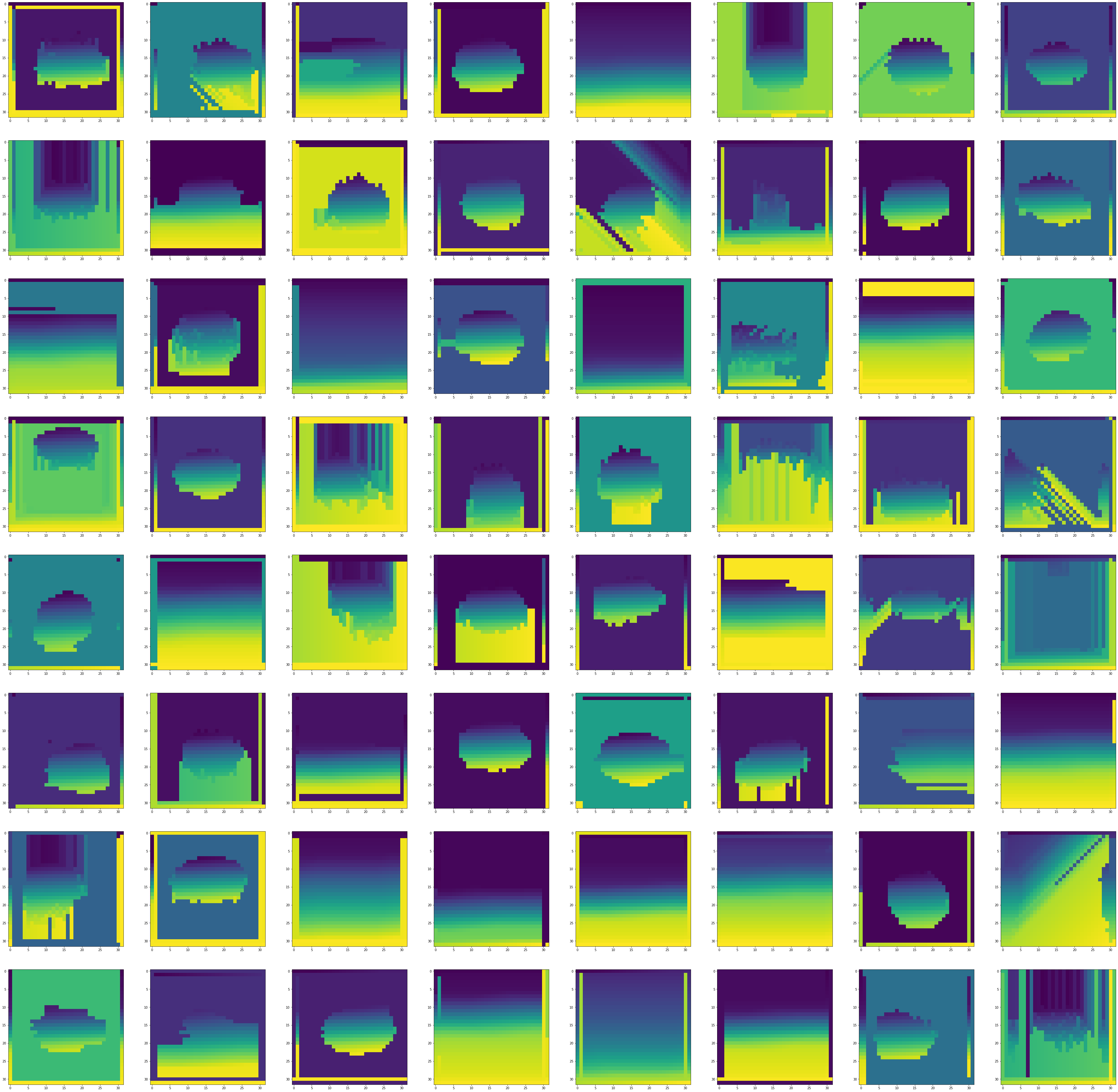}
\end{center}
  \caption{ReLU clustering. Clustering map of the 64 channels in the second layer of the CIFAR-10 network. Each channel is clustered into 256 clusters.}
\label{fig:ReLUCluster}
\end{figure}

The previous set of experiments used patches to cluster activations. Next, we evaluate the performance of clustering. To this end, we clustered activations into $k$ clusters based on the activation maps of all images in the training set. The clustering was done independently for each channel separately.

We use agglomerative clustering. First, we compute a binary activation vector per pixel, then we calculate the correlation matrix according to the hamming distance between the vectors. We initialize each activation as a set and in each iteration we take the two activations with the minimum hamming distance and if they are in different sets we unify their sets, and we stop when we have $k$ clusters. The 64 clustering maps for the 64 channels in the second layer can be seen in Figure~\ref{fig:ReLUCluster}. We can see that many channels have highly correlated background pixels that are unified into a single cluster. The clustering maps reveal information about the {\em training} examples and {\em not} about the query image.  

One can see the results of the clustering experiments in Table~\ref{table:cifar_adaptive}. In the first experiment we clustered every $8\times8$ window into $8$ clusters so in total we have $128$ clusters for each channel. In the second experiment we use image patches for the first layer where we take the middle pixel in each channel and for the second ReLU layer we use agglomerative clustering with $256$ clusters for the whole channel so in total we have the same number of ReLUs in both experiments. The results are slightly worse than the uniform $3 \times 3$ patch experiment reported earlier (See Table~\ref{table:cifar_fixed}). The results of agglomerative clustering for SVHN can be seen in table~\ref{table:svhn_fashion_fixed}.

We conclude that ReLU sharing across small image patches drops the ReLU count with minimal impact on accuracy. We further observe that agglomerative clustering does not outperform simply grouping activations into image patches. We hypothesize that this might be the case because nearby activations in the image plane are already highly correlated.

\begin{table}
\small
\begin{center}
\begin{tabular}{|l|c|c|c|c|}
\hline
Method & Layer \#1 & Layer \#2 & \#ReLU & Accuracy\\
\hline\hline
\# 1 & $(8 \times 8, 8)$ & $(8 \times 8, 8)$ & 256 & $81.96\%$\\
\# 2 & $(32 \times 32, 1)$ & $(32 \times 32, 256)$ & 257 & $81.67\%$\\
\hline
\end{tabular}
\end{center}
\caption{CIFAR-10 adaptive clustering results. The expression $(m \times m, k)$ means that we cluster every non-overlapping window of size $(m \times m$) into $k$ clusters. As can be seen, both methods use roughly the same number of ReLU operations. The first method outperforms the second one.}
\label{table:cifar_adaptive}
\end{table}

\begin{table*}
\small
\begin{center}
\begin{tabular}{|l|l|c|c|c|c|}
\hline
Dataset & Model & Accuracy & Runtime (s) & Rounds & Comm. (MB) \\
\hline\hline
CIFAR-10 & Original & $83.59\%$ & $36$ & $86$ & $141.02$ \\
& Uniform $32 \times 32$ & $81.13\%$ & $14$ & $86$ & $54.18$ \\
& FC + Noise & $82.01\%$ & $13.88$ & $90$ & $57.865$ \\
\hline
SVHN & Original & $93.14\%$ & $5.03$ & $26$ & $22.1$ \\
& Uniform $32 \times 32$ & $90.21\%$ & $1.54$ & $26$ & $5.32$ \\
\hline
Fashion-MNIST & Original & $92.11\%$ & $6.55$ & $40$ & $35.04$ \\
& Uniform $32 \times 32$ & $90.38\%$ & $3.26$ & $40$ & $19.04$ \\
\hline
\end{tabular}
\end{center}
\caption{Classification performance. We compare the performance of three different models, in terms of run time, rounds of communication, and communication bandwidth. Results are for the classification of a single image. As can be seen, for CIFAR-10, both "Uniform $32 \times 32$" and "FC + Noise" cut run time and bandwidth by about $60\%$ with little degradation in performance. For SVHN, the Uniform $32 \times 32$ model saves 70\%-76\% of the run time and communication costs and saves 45\%-50\% of the run time and communication costs in the Fashion-MNIST dataset.}
\label{table:pysyft_results2}
\end{table*}

\paragraph{From ReLU Count to Communication Bandwidth} Having established that dropping ReLU count has little impact on network accuracy, we investigate its impact on run-time, round complexity and communication bandwidth.

To do that, we evaluate 3 of the models introduced earlier that presented attractive ReLU count/accuracy trade-offs. This includes the original model, the Uniform with $32\times32$ patch size model (where we perform a single ReLU decision for each channel for the first 2 ReLU layers), and the FC + Noise model (where we perform a dot-product with a learned weight vector and add noise to the outcome). We can see the results for CIFAR-10 of classification of 1 test image in Table~\ref{table:pysyft_results2}. 

In the original model each of the 2 first ReLU layers takes 10 seconds and 46MB of communication bandwidth so just these 2 layers take over $55\%$ of the total time and over $65\%$ of the total communication bandwidth. 

In the $32\times32$ Uniform model each of the 2 first ReLU layers takes 0.25 seconds and 2.66MB, compared to 10 seconds and 46MB of the original model. This leads to  $60\%$ savings in the total time and $61\%$ savings in the total communication (the number of rounds stays the same). 

The FC + Noise model takes slightly more communication from the $32\times32$ Uniform model (but still $59\%$ less from the original model) since it has an extra fully connected layer and a multiplication with a sampled noise tensor but, as we saw in Table~\ref{table:cifar_fixed}, this model has better accuracy so we can see here a trade off between accuracy and efficiency.

We observe similar behaviour for the SVHN data set. The first ReLU layer takes 3.7 seconds and 17.87MB of communication, so over $73\%$ of the total time and over $80\%$ of total communication. In the $32\times32$ Uniform model the first ReLU layer takes 0.21 seconds and 1.09MB, 70\%-76\% less than the original model costs. 

For Fashion-MNIST, the first ReLU layer takes 3.49 seconds and 16.66MB of communication, so over $53\%$ of the total time and over $47\%$ of total communication. In the $32\times32$ Uniform model the first ReLU layer takes 0.2 seconds and 1.02MB, 45\%-50\% less than the original model costs.
We can see the results for SVHN and Fashion-MNIST of classification of 1 test image in Table~\ref{table:pysyft_results2}.

In all three models and across all three datasets we observe that drop in ReLU count translates directly into drop in communication bandwidth.

\begin{table*}
\small
\begin{center}
\begin{tabular}{|l|l|c|c|c|c|c|c|c|}
\hline
\multicolumn{2}{|c|}{\textbf{Framework}} & \textbf{Accuracy} & \multicolumn{3}{c|}{\textbf{Runtime (s)}} & \multicolumn{3}{c|}{\textbf{Comm. (MB)}} \\
\multicolumn{2}{|c|}{} & & setup & online & total & setup & online & total \\
\hline\hline
 & MiniONN & $81.61\%$ & $472$ & $72$ & $544$ & $3046$ & $6226$ & $9272$ \\
2PC & GAZELLE & $81.61\%$ & $15.5$ & $4.25$ & $19.8$ & $906$ & $372$ & $1278$ \\
 & FALCON & $81.61\%$ & $10.5$ & $3.31$ & $13.8$ & $906$ & $372$ & $1278$ \\
\hline
 & Chameleon & $81.61\%$ & $22.97$ & $29.7$ & $52.67$ & $1210$ & $1440$ & $2650$ \\
3PC & Shafran {\em et al.} & $83.53\%$ & $0$ & $19.49$ & $19.49$ & $0$ & $71.9$ & $71.9$ \\
 & OURS & $82.01\%$ & $0$ & $13.88$ & $13.88$ & $\textbf{0}$ & $\textbf{57.86}$ & $\textbf{57.86}$ \\
\hline
\end{tabular}
\end{center}
\caption{Classification performance Comparison on CIFAR-10. We compare ourselves to MiniONN~\cite{MiniONN}, GAZELLE~\cite{Gazelle}, FALCON~\cite{FALCON},  Chamelon~\cite{Chameleon}, and the crypto-oriented neural architecture design of \cite{Shafran2020}. The first three methods are 2-party MPC methods, while the last three methods take a 3-party MPC approach. In all cases our communication bandwidth is significantly lower than other methods. Our approach can be integrated into other algorithms and lead to similar savings. }
\label{table:comparison_cifar10}
\end{table*}

\paragraph{Comparison to other Approaches}
It is challenging to compare our method to other methods reported in the literature because we use different hardware and different cryptographic settings.

Specifically, we use the PySyft implementation that, in turn, relies on SecureNN~\cite{SecureNN} protocols. These are 3-party protocols, as opposed to the 2-party protocols used by others, except for Chameleon~\cite{Chameleon} that uses the third party only in the offline phase.

In our setting all 3 parties live on the same machine and do not communicate over the network, so all running times do not consider network latency. All other frameworks are evaluated in a LAN setting where each party has its own machine so their run time takes into consideration the network latency.

Previous works split their computation into an input independent offline phase and an input dependent online phase. The PySyft implementation of the secureNN protocols does not do this split and counts all cost as online cost – hence, the offline cost is 0. The online run time can be improved by performing the beaver triplet generation for multiplication in an offline phase~\cite{Beaver}.

We do observe that our work achieves a $\times 2.5$ improvement in online communication and a $\times 22$ improvement in total communication over the best results of GAZELLE~\cite{Gazelle} and FALCON~\cite{FALCON}. In online run-time our work is roughly $\times 4$ slower than the best result of FALCON~\cite{FALCON} and slightly worse in total run-time (since our run-time doesn't take network latency into consideration).

Shafran {\em et al.}~\cite{Shafran2020} used different models in their paper so in order to compare our work with theirs we applied their method on the model in table~\ref{table:cifar_cnn}. Specifically, we removed the first ReLU layer and replaced the second ReLU layer with 50\% partial activation layer. See results in table~\ref{table:pysyft_results2}. Their method does not hurt the accuracy compared to the original model, while our method requires 28\% less time and 20\% less communication bandwidth, at the cost of a roughly $1\%$ drop in accuracy.

We believe that ReLU sharing can be implemented in other frameworks and improve their run time and communication cost. For example, we can see in the benchmarks table of FALCON~\cite{FALCON} for the ReLU operation that reducing the ReLU count by an order of magnitude leads to a roughly $\times 10$ drop in both time and communication costs, for both the offline and online stages. We also observe in the FC layer benchmark there that adding another FC layer will have a negligible effect on costs. This leads us to believe that implementing ReLU sharing within the FALCON, or many other algorithms, will decrease costs.

\section{Conclusions}

We propose a method to drastically reduce the ReLU count of a CNN with minimal impact on accuracy. The proposed method groups pixels (or activation units) together and uses a single ReLU operator for all activations in the group. We discuss how to group activations and how to apply the ReLU operation.

Experiments suggest that we can reduce the ReLU count by up to three orders of magnitude which, in turn, cuts the communication bandwidth by more than $50\%$. Our approach can be used with other Privacy Preserving Machine Learning algorithms and bring us closer to practical algorithms that can be used in the wild.

\paragraph{Acknowledgments} This research was supported in part by a grant made by TATA Consultancy Services.

\bibliographystyle{aaai21}
\bibliography{egbib}

\end{document}